\pdfoutput=1
\documentclass[11pt]{article}
\usepackage[final]{acl}
\usepackage{times}
\usepackage{latexsym}
\usepackage[T1]{fontenc}
\usepackage[utf8]{inputenc}
\usepackage{microtype}
\usepackage{inconsolata}
\usepackage{graphicx}
\usepackage{booktabs}
\usepackage{multirow}
\usepackage{tabularx}
\usepackage{listings}
\usepackage{tikz}
\usetikzlibrary{arrows.meta,positioning,fit}

\newcommand{\modelopus}{\mbox{Opus-4.6}}

\title{Atomic and Holistic LLM Judges for Reference-Grounded Support Labels:\\
A Prompt-Controlled Comparison}

\author{Xinran Zhang \\
  University of California, Berkeley \\
  \texttt{zhangxr7@berkeley.edu}}

\hypersetup{pdftitle={Atomic and Holistic LLM Judges for Reference-Grounded Support Labels: A Prompt-Controlled Comparison}, pdfauthor={Xinran Zhang}}

\begin{document}
\maketitle

\begin{abstract}
When an LLM judge only has to assign a three-way support label to a candidate answer given a reference, does asking it to decompose the answer into atomic claims help, and at what cost? We compare four single-call designs that share the judge model, the inputs, and the level of instruction detail: candidate-side atomic decomposition, a matched holistic rubric, reference-side decomposition that checks whether each reference claim is covered, and a bidirectional combination. Support labels are constructed from TruthfulQA, ASQA, and QAMPARI references (200 questions and 400 rows per dataset). All four designs are run with \modelopus{}, GPT-4.1, and Gemini Flash Lite; Sonnet-4.6 is added for the candidate-side and holistic designs. Candidate-side decomposition is weak where the label depends on completeness: the holistic rubric is more accurate on ASQA and QAMPARI for every judge while using fewer tokens. Reference-side decomposition is 12.5--21.3 points more accurate than holistic on ASQA for 28--29\% more tokens, and stays near the holistic ceiling on QAMPARI at 63--69\% more. On TruthfulQA misconceptions, candidate-side decomposition is competitive and significantly better for two judges. On a 60-row single-author subset whose labels are looser than strict reference completeness, candidate-side decomposition edges ahead of holistic for every judge, reversing the construction-label order. What a judge decomposes should follow what the label measures.
\end{abstract}

\section{Introduction}
Reference-grounded LLM judges are routinely asked for a single label: is a candidate answer supported by a reference, partially supported, or unsupported? A popular way to make such judgments more careful is atomic decomposition. FActScore \citep{min2023factscore} split generations into atomic facts and verified each one, and the idea now runs through long-form factuality pipelines \citep{wei2024long, liu-etal-2025-verifact, wan-etal-2025-fastfact}. In a single judging call, the common instantiation decomposes the \emph{candidate}: list its claims, check each against the reference, and aggregate.

A support label, however, mixes properties that decomposition treats differently. We distinguish three. \emph{Factual precision} asks whether every claim the candidate makes is backed by the reference. \emph{Strict reference completeness} asks whether the candidate covers what the reference contains. \emph{Claim-level artifacts}---claim inventories, per-claim decisions, audit trails---are useful products in their own right. Candidate-side decomposition targets precision and produces artifacts, but nothing in its structure lists what the candidate leaves out. When the label rewards completeness, a judge can verify every claim an incomplete answer makes and still call it supported.

We measure this in a prompt-controlled comparison of four single-call designs (Figure~\ref{fig:designs}): candidate-side atomic (C), a matched holistic rubric (H), reference-side atomic (R), which decomposes the reference and checks coverage, and bidirectional atomic (B), which does both. Within each comparison the designs share the judge model, the inputs, and the level of instruction detail, but not realized computation; token cost is reported as an outcome (Section~\ref{sec:prompt-control}). We score the designs against support labels constructed from TruthfulQA \citep{lin2022truthfulqa}, ASQA \citep{stelmakh2022asqa}, and QAMPARI \citep{yoran2022qampari}, with 200 questions (400 rows) per dataset, three judge families, and Sonnet-4.6 as a fourth judge for C and H.

The direction of decomposition matters more than its presence. Candidate-side decomposition is the weakest design on the two completeness-heavy datasets. On ASQA its accuracy is 7.25--15.5 points below the holistic rubric for all four judges, and GPT-4.1 and Gemini Flash Lite recognize only 16\% and 12.5\% of partially supported rows as partial. Reference-side decomposition reverses this: it is the most accurate design on ASQA for all three judges on which it was run (.8675--.975, sign-test \(p<10^{-14}\) against holistic) at about 28--29\% more tokens than holistic, and on QAMPARI it stays near the holistic ceiling while costing 63--69\% more. TruthfulQA, whose negative candidates are misconceptions that contradict the reference, points the other way. There, candidate-side decomposition is at least as accurate as the rubric for every judge, significantly so for Gemini Flash Lite and Sonnet-4.6, and slightly cheaper.

These are measurements against constructed labels, and we keep that scope explicit. Supported candidates are verbatim copies of the reference, and ASQA/QAMPARI partials are lexical halves of the benchmark answer sets. These partial labels encode strict reference completeness, not necessarily factual error. Two smaller checks probe that choice without replacing it. On a 60-row subset labeled by one author, the author's labels are looser than the ASQA construction labels. Under them, the candidate-side judge is ahead of the holistic judge for every judge model, while the construction labels put holistic ahead (Section~\ref{sec:human-sanity}). The second check uses LLM-generated ASQA answers. There, nominal full answers, which are not verified to be complete, are paired with mechanically truncated partials, and both sets of construction labels are noisy. Measured against these labels, holistic still agrees more often than candidate-side decomposition, and the reference-side design flags most nominal full answers as incomplete (Section~\ref{sec:llmgen}). Neither check establishes performance against independently validated factuality labels. Together they show that the preferred design depends on the standard a label encodes.

Our contribution is measurement and design guidance rather than a new judge: (i) a prompt-controlled, cost-accounted comparison of four decomposition designs across four judges; (ii) source-level evidence that candidate-side and reference-side decomposition behave differently on completeness-heavy labels; and (iii) diagnostics and label-semantics checks that separate the standard a label encodes from the judge scored against it.

\section{Related Work}
\paragraph{Claim-level factuality evaluation.}
FActScore \citep{min2023factscore} evaluates long-form generations by splitting them into atomic facts and verifying each one against a knowledge source, and SAFE \citep{wei2024long} pairs multi-step decomposition with search-based verification. VeriFact \citep{liu-etal-2025-verifact} refines fact extraction and builds reference fact sets for long responses, which adds a recall view close to our reference-side design. FaStFACT \citep{wan-etal-2025-fastfact} speeds up such pipelines and strengthens evidence collection, and UFO \citep{huang-etal-2024-ufo} shows that verdicts depend on which fact source the evaluator receives. Verification signals also steer generation. A self-correcting RAG framework, for example, combines knapsack-style context selection with NLI-guided tree search to keep answers faithful to the retrieved evidence \citep{xu2026self}. These systems usually run over several calls with retrieval, and they mostly target precision. We study a single evaluator call with a supplied reference and ask which side of the comparison should be decomposed.

\paragraph{Reference-guided and rubric-based judges.}
Explicit references improve LLM evaluation of free-form QA \citep{badshah2025reference}, and YESciEval \citep{dsouza-etal-2025-yescieval} develops robust rubric-based LLM judging for scientific QA. F\textsuperscript{2}Bench \citep{lan-etal-2025-f2bench} scores open-ended responses against given facts with a rubric covering accurate reflection of those facts, unwarranted generalization, and unsupported claims, a supplied-evidence design close in form to our holistic judge. A reference also changes where the difficulty lies. Closed-book hallucination has been analyzed in terms of what a finite parametric memory can retain of its training facts \citep{wang2026costcompressionratedistortionlimit}. When the evidence is in the input instead, the task is to map the candidate--reference relation onto the label definition, including what counts as complete. Prior reference-guided work motivates this setting, but it leaves decomposition direction, instruction detail, and cost entangled.

\paragraph{Reliability of LLM judges.}
Frontier judges generalize better than fine-tuned substitutes \citep{huang-etal-2025-empirical}, judge behavior varies across languages \citep{fu-liu-2025-reliable}, repeated ratings can be self-inconsistent \citep{haldar-hockenmaier-2025-rating}, and replacing human annotators with judges calls for proper statistical testing \citep{calderon-etal-2025-alternative}. Supplied evidence does not automatically override model-internal knowledge \citep{qi-etal-2025-evaluating}, and QA judges can fail when references conflict with what the model already knows \citep{lee-etal-2026-judging}. Models can also read past the context they are given. On the context-flip PaCE benchmark they prefer familiar pragmatic readings over the literal reading the context requires, and explicit strict-literal instructions change the outcome \citep{li-etal-2026-born}. Task format matters as well: forcing multiple-choice answers changes the safety behavior of 14 models \citep{chen2026choicesriskssafetyfailures}. We therefore treat the judging instruction as an experimental variable and measure it together with its token cost.

\section{Task, Designs, and Protocol}

\subsection{Support labels constructed from benchmarks}
\label{sec:labels}
Given a question, a reference answer, and a candidate answer, we ask the judge to return one JSON object whose \texttt{verdict} is \texttt{supported}, \texttt{partially\_supported}, or \texttt{unsupported}. Omissions count: a candidate that is correct as far as it goes but leaves out reference content is partially supported.

The benchmarks do not provide such three-way labels; we construct them from benchmark references and candidates (Table~\ref{tab:dataset-classes}; Appendix~\ref{app:dataset}). Each of 200 source questions per dataset yields two rows. In all three datasets the supported candidate is the reference text itself, verbatim. On TruthfulQA the other candidate is an incorrect answer from the benchmark, typically a common misconception, labeled unsupported. On ASQA the partial candidate renders the first half of the benchmark’s disambiguated \texttt{qa\_pairs} as ‘sub-question + first short answer’ fragments; on QAMPARI it lists the first half of the gold answer list. Both partial constructions are deterministic and lexical.

Three consequences follow. Supported rows mainly test whether a judge accepts an exact copy of the reference, and every design scores .94--1.00 on them; the discriminating rows are the partial and unsupported ones. Partial labels encode coverage of the reference, not factual error: an ASQA partial row can be correct in every claim it makes. We therefore call them \emph{construction labels} and read agreement with them as agreement with a strict reference-completeness standard; Section~\ref{sec:human-sanity} examines a looser standard. Finally, only the \texttt{verdict} field is scored. Scores, confidences, rationales, and claim lists do not enter accuracy.

\begin{table}[t]
\centering
\small
\begin{tabular}{lrrr}
\toprule
Dataset & \texttt{supp.} & \texttt{partial} & \texttt{unsupp.} \\
\midrule
TruthfulQA & 200 & --- & 200 \\
ASQA       & 200 & 200 & --- \\
QAMPARI    & 200 & 200 & --- \\
\bottomrule
\end{tabular}
\caption{Construction-label composition of the 400 rows per dataset (two rows for each of 200 questions). Supported candidates are verbatim copies of the reference; the \texttt{partially\_supported} rows of ASQA and QAMPARI carry the completeness signal.}
\label{tab:dataset-classes}
\end{table}

\subsection{Judge designs}
All designs present the question, reference answer, and candidate answer in the same user-message layout, followed by the requested JSON keys. The first two keys are \texttt{verdict} and \texttt{score}; the system instruction and remaining keys depend on the design (Appendix~\ref{app:prompt-templates}).

\paragraph{C: candidate-side atomic.} The judge breaks the candidate into atomic claims and grounds each claim in the reference (keys \texttt{atomic\_claims}, \texttt{supported\_claims}, \texttt{unsupported\_claims}). Its main prompt (v1) does not mention completeness.

\paragraph{H: matched holistic.} The judge evaluates the candidate with a rubric that lists correctness, completeness, unsupported detail, and robustness to style bias, and is told not to decompose the answer explicitly (keys \texttt{evidence\_summary}, \texttt{unsupported\_or\_missing\_points}).

\paragraph{R: reference-side atomic.} The judge breaks the reference into atomic claims and checks whether the candidate covers each one (\texttt{reference\_claims}, \texttt{covered\_claims}, \texttt{missing\_claims}). It is not asked about candidate content outside the reference.

\paragraph{B: bidirectional atomic.} The judge does both and returns all six claim lists.

\smallskip\noindent
A schema-matched variant (C’s keys without the decomposition instruction) and a holistic pairwise judge appear only in the diagnostics.

\begin{figure}[t]
\centering
\begin{tikzpicture}[
  font=\scriptsize,
  io/.style={draw, rounded corners=2pt, fill=gray!12, align=center, text width=7.0cm, inner sep=3pt},
  des/.style={draw, rounded corners=2pt, align=center, text width=3.2cm, minimum height=15mm, inner sep=3pt},
  arr/.style={-{Stealth[length=1.6mm]}, semithick}
]
\node[io] (in) at (0,0) {\textbf{Shared input:} question, reference \(r\), candidate \(c\)\\ \textbf{Requested output:} one JSON object, \texttt{verdict} first};
\node[des] (C) at (-1.8,-1.55) {\textbf{C: candidate-side}\\ split \(c\) into claims;\\ check each claim in \(r\)\\ \emph{targets precision}};
\node[des] (H) at (1.8,-1.55) {\textbf{H: matched holistic}\\ rubric incl.\ completeness;\\ no explicit decomposition\\ \emph{targets both, implicitly}};
\node[des] (R) at (-1.8,-3.35) {\textbf{R: reference-side}\\ split \(r\) into claims;\\ check each is covered by \(c\)\\ \emph{targets completeness}};
\node[des] (B) at (1.8,-3.35) {\textbf{B: bidirectional}\\ split both \(c\) and \(r\);\\ check support and coverage\\ \emph{targets both, explicitly}};
\node[io] (out) at (0,-4.95) {\textbf{Scored:} \texttt{verdict} vs.\ construction label; tokens per call};
\node[draw, rounded corners=2pt, fit=(C)(H)(R)(B), inner sep=2mm] (designs) {};
\draw[arr] (in.south) -- (designs.north);
\draw[arr] (designs.south) -- (out.north);
\end{tikzpicture}
\caption{Four alternative single-call judge designs. Each receives the same inputs and returns a verdict; the instructions differ in what the judge decomposes and checks. One design is used per call, not a sequence of four stages. Prompts are in Appendix~\ref{app:prompt-templates}.}
\label{fig:designs}
\end{figure}

\subsection{What ‘prompt-controlled’ means}
\label{sec:prompt-control}
We call the comparison \emph{prompt-controlled}: within each comparison, the designs use the same judge model, the same inputs in the same layout, and instructions of comparable detail. It is not compute-matched. Output length is not fixed, the designs write different amounts, and realized tokens are part of the outcome; any accuracy gain is assessed alongside its token cost. Prompt control has two further limits. An instruction not to decompose constrains what the judge writes, not what it computes, so nothing here is a claim about internal mechanisms. And the prompts are not symmetric in what they make explicit: the holistic rubric names completeness, while candidate-side v1 does not. Candidate-side v2, which asks for ‘factual accuracy and completeness’, is one check on this asymmetry (Section~\ref{sec:diagnostics}). Our inferences are scoped to this single-call prompting protocol.

\subsection{Judges, scoring, and statistics}
The judges are \modelopus{}, GPT-4.1, and Gemini Flash Lite for all four designs, and Sonnet-4.6 for C and H. Gemini Flash Lite supplies a third model family and a lightweight model tier, not a calibrated measure of judge capability. Logged identifiers and diagnostic-specific settings are listed in Appendix~\ref{app:reproducibility}. The primary runs specify neither temperature nor an output-token limit; the actual provider-default values were not recorded. We report accuracy over the 400 rows, class-wise accuracy, and mean total (input plus output) tokens per call. Output-token distributions are reported where available (Appendix~\ref{app:stats}); the Gemini interface recorded total but not output tokens.

Our inferential comparison is an exact two-sided sign test at the source-question level. For each of the 200 questions we count the rows (0--2) each design labels correctly, exclude ties, and test whether the remaining questions favor either design with equal probability. Reported \(p\)-values are unadjusted for multiple comparisons. For Opus and GPT-4.1, we also report source-clustered bootstrap intervals and stability under sampling half of the questions without replacement (Appendix~\ref{app:stats}). Figure~\ref{fig:flow} summarizes the protocol.

\begin{figure}[t]
\centering
\begin{tikzpicture}[
  font=\scriptsize,
  box/.style={draw, rounded corners=2pt, align=center, text width=3.65cm, inner sep=3pt},
  side/.style={draw, dashed, rounded corners=2pt, align=center, text width=2.75cm, inner sep=3pt},
  arr/.style={-{Stealth[length=1.6mm]}, semithick}
]
\node[box, fill=gray!12] (b1) {Benchmark question + reference\\ TruthfulQA, ASQA, QAMPARI};
\node[box, below=3mm of b1] (b2) {Two rows per question\\ supported: \(c = r\) verbatim\\ partial: first half of answers\\ unsupported: incorrect answer};
\node[box, below=3mm of b2] (b3) {One judge call per row\\ design \(\in\) \{C, H, R, B\} \(\times\) judge};
\node[box, below=3mm of b3] (b4) {Recover the \texttt{verdict} from the output};
\node[box, fill=gray!12, below=3mm of b4] (b5) {Accuracy vs.\ construction labels; source-level sign tests; tokens};
\node[side, right=3.5mm of b2] (s1) {\textbf{Diagnostics} (\S\ref{sec:diagnostics})\\ wording, schema, budget, reference quality, perturbations};
\node[side, right=3.5mm of b4] (s2) {\textbf{Scope checks}\\ generated answers (\S\ref{sec:llmgen});\\ author labels (\S\ref{sec:human-sanity})};
\draw[arr] (b1) -- (b2);
\draw[arr] (b2) -- (b3);
\draw[arr] (b3) -- (b4);
\draw[arr] (b4) -- (b5);
\draw[arr, dashed] (b2) -- (s1);
\draw[arr, dashed] (s2) -- (b4);
\end{tikzpicture}
\caption{Evaluation flow. Primary runs use 200 questions (400 rows) per dataset. Controlled diagnostics use 40--50-row slices; generated candidates and a 60-row author-annotated slice probe the scope of the construction-label results.}
\label{fig:flow}
\end{figure}

\section{Results Against Construction Labels}
\label{sec:results}
Table~\ref{tab:main-results} gives accuracy and mean total tokens for every evaluated dataset, design, and judge combination. Appendix~\ref{app:stats} gives supporting statistics and complexity bins.

\begin{table*}[t]
\centering
\small
\setlength{\tabcolsep}{5pt}
\begin{tabular}{llrrrrrrrr}
\toprule
\multirow{2}{*}{Dataset} & \multirow{2}{*}{Design} & \multicolumn{2}{c}{\modelopus{}} & \multicolumn{2}{c}{GPT-4.1} & \multicolumn{2}{c}{Gemini Flash Lite} & \multicolumn{2}{c}{Sonnet-4.6} \\
\cmidrule(lr){3-4} \cmidrule(lr){5-6} \cmidrule(lr){7-8} \cmidrule(lr){9-10}
 & & Acc. & Tokens & Acc. & Tokens & Acc. & Tokens & Acc. & Tokens \\
\midrule
\multirow{4}{*}{TruthfulQA} & C & 0.9750 & 336.40 & 0.9775 & 284.52 & 0.9625 & 295.27 & \textbf{0.9750} & 353.84 \\
 & H & 0.9700 & 364.66 & 0.9675 & 292.44 & 0.9300 & 309.33 & 0.9450 & 381.79 \\
 & R & \textbf{0.9850} & 326.93 & 0.9725 & 285.87 & 0.9550 & 295.53 & --- & --- \\
 & B & 0.9800 & 430.93 & \textbf{0.9800} & 378.27 & \textbf{0.9675} & 397.03 & --- & --- \\
\midrule
\multirow{4}{*}{ASQA} & C & 0.7775 & 569.92 & 0.5650 & 475.19 & 0.5500 & 504.52 & 0.7150 & 588.09 \\
 & H & 0.8500 & 492.81 & 0.7200 & 410.12 & 0.6975 & 423.74 & \textbf{0.8100} & 507.41 \\
 & R & \textbf{0.9750} & 631.20 & \textbf{0.9325} & 526.37 & \textbf{0.8675} & 546.97 & --- & --- \\
 & B & 0.9000 & 852.08 & 0.8025 & 700.80 & 0.7550 & 737.85 & --- & --- \\
\midrule
\multirow{4}{*}{QAMPARI} & C & 0.9750 & 803.47 & 0.9275 & 676.45 & 0.9525 & 599.20 & 0.9625 & 785.66 \\
 & H & 0.9925 & 510.68 & \textbf{0.9975} & 420.06 & 0.9925 & 423.29 & \textbf{0.9950} & 521.22 \\
 & R & \textbf{0.9950} & 862.14 & 0.9950 & 701.21 & \textbf{0.9975} & 688.87 & --- & --- \\
 & B & 0.9825 & 1258.16 & 0.9875 & 981.46 & 0.9925 & 895.60 & --- & --- \\
\bottomrule
\end{tabular}
\caption{Accuracy against construction labels and mean total tokens per call, 400 rows (200 questions) per cell. C = candidate-side atomic, H = matched holistic, R = reference-side atomic, B = bidirectional atomic. Bold marks the most accurate design per dataset and judge. Sonnet-4.6 was run with C and H only.}
\label{tab:main-results}
\end{table*}

\subsection{Candidate-side decomposition versus the holistic rubric}
On ASQA the holistic rubric is more accurate than candidate-side decomposition for all four judges, by 7.25 to 15.5 points, and uses 65--81 fewer tokens per call (Table~\ref{tab:significance}; source-level \(p\le 3\times10^{-8}\)). The gap sits in the partial rows, where holistic minus candidate-side accuracy is +14.5, +30.5, +33.0, and +19.0 points for \modelopus{}, GPT-4.1, Gemini Flash Lite, and Sonnet-4.6 (Table~\ref{tab:class}). QAMPARI shows the same ordering with smaller margins: holistic is 1.75--7.0 points more accurate (\(p\le .016\)) and 176--293 tokens cheaper, because candidate-side outputs enumerate long answer lists (median output 269--348 tokens, against 118--161 for holistic; Table~\ref{tab:lengths}).

TruthfulQA reverses the sign. Candidate-side decomposition is at least as accurate as the rubric for every judge, and the difference is significant for Gemini Flash Lite and Sonnet-4.6 (12:0 questions each, \(p=4.9\times10^{-4}\)); for \modelopus{} and GPT-4.1 it rests on four discordant questions and is not significant. The edge comes mainly from unsupported rows (holistic minus candidate-side: \(-1.0\) to \(-6.0\) points), and candidate-side decomposition is also 8--28 tokens cheaper on this dataset. Appendix~\ref{app:qualitative} gives complete recorded examples and a JSON verdict.

The source-clustered intervals in Table~\ref{tab:bootstrap} preserve each question's two-row dependence. For ASQA and QAMPARI, at least 98.8\% of the half-question subsets preserve the direction of the full-data accuracy difference. The bootstrap and sign test are not interchangeable: for TruthfulQA GPT-4.1, the percentile interval excludes zero while the exact sign-test \(p=.125\).

\begin{table}[t]
\centering
\small
\setlength{\tabcolsep}{4pt}
\begin{tabular}{lrrrr}
\toprule
Judge & \(\Delta\)Acc. & \(\Delta\)Tokens & H:C & \(p\) \\
\midrule
\multicolumn{5}{l}{\emph{TruthfulQA}} \\
\modelopus{} & \(-\)0.0050 & +28.27 & 1:3 & 0.6250 \\
GPT-4.1 & \(-\)0.0100 & +7.92 & 0:4 & 0.1250 \\
Gemini FL & \(-\)0.0325 & +14.06 & 0:12 & 4.88e-04 \\
Sonnet-4.6 & \(-\)0.0300 & +27.95 & 0:12 & 4.88e-04 \\
\midrule
\multicolumn{5}{l}{\emph{ASQA}} \\
\modelopus{} & +0.0725 & \(-\)77.11 & 30:1 & 2.98e-08 \\
GPT-4.1 & +0.1550 & \(-\)65.07 & 66:5 & 1.19e-14 \\
Gemini FL & +0.1475 & \(-\)80.78 & 67:8 & 1.01e-12 \\
Sonnet-4.6 & +0.0950 & \(-\)80.68 & 41:3 & 1.62e-09 \\
\midrule
\multicolumn{5}{l}{\emph{QAMPARI}} \\
\modelopus{} & +0.0175 & \(-\)292.79 & 7:0 & 0.0156 \\
GPT-4.1 & +0.0700 & \(-\)256.39 & 28:0 & 7.45e-09 \\
Gemini FL & +0.0400 & \(-\)175.91 & 18:2 & 4.02e-04 \\
Sonnet-4.6 & +0.0325 & \(-\)264.44 & 13:0 & 2.44e-04 \\
\bottomrule
\end{tabular}
\caption{Holistic (H) minus candidate-side (C). \(\Delta\)Acc.\ \(>0\) favors H; \(\Delta\)Tokens \(<0\) means H is cheaper. H:C counts questions (of 200) on which H or C labels more rows correctly; \(p\) is the unadjusted exact two-sided sign test after excluding tied questions.}
\label{tab:significance}
\end{table}

\subsection{Reference-side and bidirectional decomposition}
Reference-side decomposition is the most accurate design on ASQA for every judge on which it was run: .975, .9325, and .8675 for \modelopus{}, GPT-4.1, and Gemini Flash Lite. That is 12.5, 21.25, and 17.0 points above holistic and 19.75--36.75 points above candidate-side decomposition. Against holistic it wins on 50:0, 84:1, and 69:2 questions (\(p\le 1.8\times10^{-15}\); Table~\ref{tab:ref-sign}). Its partial-row accuracy is .95, .87, and .745, against .555, .16, and .125 for the candidate-side design, so the same judges on the same inputs separate complete from incomplete answers far better when the decomposition runs over the reference. The price is about 28--29\% more tokens than holistic and 8--11\% more than candidate-side decomposition.

On QAMPARI, reference-side accuracy (.995--.9975) is close to the holistic ceiling (.9925--.9975; 1:0, 1:2, and 2:0 questions) at 63--69\% more tokens, and it improves on candidate-side decomposition for all three judges (\(p\le .008\)). On TruthfulQA it stays within 2.5 points of the other designs. It beats holistic for \modelopus{} (6:0, \(p=.031\)) and Gemini Flash Lite (11:2, \(p=.022\)) and does not differ significantly from candidate-side decomposition.

Bidirectional decomposition does not simply combine the two strengths. On ASQA it improves on holistic (.900, .8025, .755; 23:3, 38:6, and 30:8 questions, \(p\le 4.72\times10^{-4}\)), but it loses to reference-side decomposition on 30--54 questions while winning on at most 5, and it costs 71--74\% more tokens than holistic. Its partial-row accuracy (.80, .61, .52) lies between the candidate-side and reference-side designs. On QAMPARI it is no better than holistic at 2.1--2.5 times the tokens; on TruthfulQA it is the most accurate design for GPT-4.1 and Gemini Flash Lite, but it beats holistic significantly only for Gemini (14:0).

The ASQA partial label states, by construction, that half of the reference’s disambiguated answers are missing, and a reference-side judge performs almost exactly that check. Its near-ceiling agreement is therefore strong evidence that it enforces strict reference completeness. It is not by itself evidence of better factuality judgment, and Section~\ref{sec:human-sanity} shows that the distinction matters.

\subsection{Complexity and a fourth judge}
\paragraph{Complexity bins.} We stratify by the number of ASQA \texttt{qa\_pairs} (2, 3, 4, 5+) and of QAMPARI answers (5, 6--7, 8--9, 10+), which are reference-complexity proxies rather than measured atomic-claim counts. Across the three primary judges, holistic minus candidate-side accuracy is positive in 21 of 24 bin--judge cells and zero in the other three, all on QAMPARI (Table~\ref{tab:bins}). There is no monotone trend on ASQA; on QAMPARI the GPT-4.1 and Gemini gaps are larger for questions with eight or more answers (up to +12.3 points). The ordering is not confined to one inspected complexity bin, but these descriptive strata do not establish a universal complexity effect.

\paragraph{Sonnet-4.6.} Adding Sonnet-4.6 for C and H reproduces the pattern with a different judge. Holistic is better on ASQA (.810 vs.\ .715; 41:3 questions) and QAMPARI (.995 vs.\ .9625; 13:0), and candidate-side decomposition is better on TruthfulQA (.975 vs.\ .945; 12:0 in its favor). The TruthfulQA edge is therefore not confined to the weakest judge. One account is that claim-by-claim checking acts as a scaffold for detecting contradictions, which TruthfulQA’s misconception candidates require, while it offers no inventory for detecting omissions. We present this as a hypothesis: the designs differ in more than decomposition, and we ran no matched comparison of reasoning and non-reasoning model variants.

\subsection{Prompt variant stability}
\label{sec:prompt-variants}
The C--H contrast could depend on the wording of one prompt pair. To check this, we wrote two more candidate-side and two more holistic templates (v2, v3). Each keeps its design's structure but changes the framing, step order, and emphasis (Appendix~\ref{app:prompt-templates}). We ran them on diagnostic slices (Table~\ref{tab:prompt-variants}). On ASQA, every holistic variant is more accurate than every candidate-side variant, both for \modelopus{} (.88--.94 vs.\ .78) and for Sonnet-4.6 (.88 vs.\ .72--.76). Candidate-side v2 explicitly asks for ``factual accuracy and completeness'' and still does not close the gap. The gap persists even with that explicit completeness instruction. On TruthfulQA, all six \modelopus{} variants fall within .95--.975. The alternative templates are longer and cost more tokens. Because the slices are small, we read the ASQA results as evidence for a consistent ordering under these wordings, not a stable effect size.

\section{Diagnostics}
\label{sec:diagnostics}
The checks below compare C and H on slices of 40--50 rows, mostly with \modelopus{} and GPT-4.1; setup details and full tables are in Appendices~\ref{app:diagnostics} and~\ref{app:robustness}. They test possible explanations for the ASQA gap but do not identify a mechanism.

\paragraph{Output schema.} We ran a schema-matched judge that returns C's claim-list keys without the decomposition instruction. On a 50-row ASQA slice it reaches .800 (\modelopus{}) and .620 (GPT-4.1), below the holistic accuracies on the same rows (Table~\ref{tab:schema-control}). On TruthfulQA it matches the separate formal C diagnostic runs (.950 for \modelopus{} and .975 for GPT-4.1) at a higher token cost. Matching the candidate-side output keys without requesting decomposition does not close the gap with H.

\paragraph{Output budget.} With explicit output caps of 128, 256, and 512 tokens, the ASQA ordering is the same for both judges: .76 for C against .86--.88 for H with \modelopus{}, and .58--.60 against .66--.70 with GPT-4.1. TruthfulQA stays essentially flat (Table~\ref{tab:budget-control}). The ASQA C--H ordering persists across the three tested output caps.

\paragraph{Reference quality.} Replacing the reference with a deterministic truncation leaves TruthfulQA accuracy unchanged, and removing the reference entirely lowers it to .55--.75. On ASQA, truncation alone costs 20--24 points for both designs and both judges (Table~\ref{tab:grounding-control}).

\paragraph{Perturbations.} Substituting another example's reference flips 34--50\% of ASQA verdicts. Appending a fixed distractor notice or stylistic padding flips at most 10\%, and TruthfulQA flip rates stay at or below 5\% (Table~\ref{tab:robustness}). Reversing candidate order in the pairwise holistic judge changes at most 4\% of verdicts.

\smallskip\noindent
Taken together, the ASQA verdicts depend heavily on the content of the reference, and the C--H gap persists across wording changes, the schema-matched control, and output caps. These are sensitivities measured on slices, not causal evidence about what either design computes.

\section{Checks on Label Semantics}

\subsection{Author labels on a 60-row subset}
\label{sec:human-sanity}
The author labeled 60 rows with the three-way scheme. They come from 30 source questions: ten two-candidate pairs per dataset. Author and construction labels agree on 51 of the 60 rows, including all 20 TruthfulQA rows, all 20 QAMPARI rows, and 11 of the 20 ASQA rows (Table~\ref{tab:human-sanity}). All nine disagreements are ASQA construction-partial rows that the author labeled supported. The author treated these candidates as factually correct but not exhaustive. For example, ``December 6, 1865'' is given as the ratification date of the 13th Amendment, while the reference also covers the amendment's passage. The author's standard is thus closer to factual correctness than to strict reference completeness. Appendix~\ref{app:human-annotations} lists the disagreement cases.

\begin{table}[t]
\centering
\small
\setlength{\tabcolsep}{5pt}
\begin{tabular}{lrrrrrr}
\toprule
 & \multicolumn{3}{c}{Construction} & \multicolumn{3}{c}{Author} \\
\cmidrule(lr){2-4} \cmidrule(lr){5-7}
Judge & C & H & R & C & H & R \\
\midrule
\modelopus{} & 53 & 55 & \textbf{59} & \textbf{54} & 52 & 50 \\
GPT-4.1 & 49 & 50 & \textbf{56} & \textbf{54} & 53 & 51 \\
Gemini FL & 48 & 50 & \textbf{55} & \textbf{55} & 53 & 52 \\
\bottomrule
\end{tabular}
\caption{Correct verdicts out of 60 rows (30 questions, ten pairs per dataset), scored against construction labels and against a single author's labels. The two label sets differ on nine ASQA rows. The counts are descriptive.}
\label{tab:human-judges}
\end{table}

The two standards rank the designs differently (Table~\ref{tab:human-judges}). Under construction labels, the order matches the primary evaluation for every judge, with R ahead of H and H ahead of C. Under the author's labels, C is ahead of H for every judge (54 vs.\ 52, 54 vs.\ 53, 55 vs.\ 53), and R is last for every judge. The reversal follows from the nine disagreement rows. On them, the author's labels reward the \texttt{supported} verdict that the construction labels count as an error, so the designs that best detect omitted reference content lose the most. The C--H differences are one or two rows. The labels come from one author, with no inter-annotator agreement, and we do not verify that the author was blind to judge outputs. The subset therefore shows that the preferred design depends on the standard a label encodes. It does not estimate how often human annotators in general would prefer one design.

\subsection{Generated ASQA answers}
\label{sec:llmgen}
To test candidates that were not cut from benchmark answers, we had \modelopus{} and Sonnet-4.6 each answer the same 50 ASQA questions, which gives 100 generated answers. Each answer yields two rows, 200 in total. The full answer is labeled \texttt{supported}. A nominal partial is built by splitting the answer on ``.\ '', keeping the first third of the segments, and appending a period; it is labeled \texttt{partially\_supported}. Both sets of construction labels are noisy. A fluent generated answer need not cover the ASQA reference, so some ``full'' rows are arguably partial. The splitter is also not sentence-aware: 25 of the 100 nominal partials (10 from \modelopus{}, 15 from Sonnet-4.6) are the full answer with a period appended. We report agreement with these construction labels only.

Pooled over the 200 rows, H agrees with the labels more often than C for every judge: 95 vs.\ 88 for \modelopus{}, 113 vs.\ 76 for GPT-4.1, and 123 vs.\ 103 for Gemini Flash Lite. H's overall accuracy is at least C's for both generators and all three judges (Tables~\ref{tab:llmgen} and~\ref{tab:llmgen-split}). R labels only 16--38\% of full rows correctly, below both C and H in every judge--generator cell. It scores 74--78\% on nominal partials, but that figure includes rows whose content is identical to their full counterparts, so it mostly reflects R's tendency to flag generated answers as incomplete. \modelopus{} judges its own generations in half of its rows. Disaggregating by generator makes this overlap visible but cannot rule out self-preference. Per-call decoding settings were not retained for answer generation; the generation code defaults to temperature 0.7 and a 256-token cap. Among the judging runs, only the \modelopus{} C and H calls set a 256-token output cap (Appendix~\ref{app:reproducibility}). We treat this set as a noisy stress test. It does not measure free-form factuality accuracy.

\section{Discussion}
The results do not support a single default for decomposition. They support a rule for choosing what to decompose. When a label rewards coverage of the reference, as the ASQA and QAMPARI construction labels do, candidate-side decomposition has no inventory of what is missing. The ASQA gap persists when C explicitly names completeness, in the schema-matched control, and under all three output caps. Reference-side decomposition supplies the missing inventory and is the most accurate design on ASQA at moderate extra cost. The matched holistic rubric is a cheaper and strong baseline that should be reported before paying for decomposition. When the label depends on contradiction, as with TruthfulQA misconceptions, candidate-side checking is competitive and sometimes better.

The scope checks show why the target label matters. Reference-side decomposition agrees well with strict completeness labels but less well with the author's labels. The author's labels reverse the primary ranking; on generated answers, R's advantage over H does not persist across judges. Practitioners should therefore decide first whether a label means factual correctness or coverage of a particular reference, and check that the reference is complete and appropriate for that meaning. Only then should they choose which side to decompose. Claim-level artifacts remain a separate reason to decompose. Their value should be weighed against their token cost and should not be assumed to raise label accuracy.

\section{Conclusion}
We compared candidate-side, holistic, reference-side, and bidirectional single-call judges under matched inputs and instruction detail, with token cost reported as an outcome. Against strict reference-completeness labels, candidate-side decomposition is the weakest design on ASQA and QAMPARI. Reference-side decomposition is the strongest on ASQA, and the holistic rubric sits between them at the lowest cost on that dataset. On TruthfulQA misconceptions, candidate-side decomposition is competitive or better. A small single-author check shows that rankings depend on label semantics; the noisy generated-answer set leaves transfer to independently validated answers unresolved. What a judge decomposes should follow what the label measures.

\section*{Limitations}
\label{sec:limitations}
Our labels are constructed. Supported candidates are verbatim references, and ASQA and QAMPARI partials are deterministic lexical halves of the benchmark answer sets, so the primary accuracies measure agreement with a strict reference-completeness standard in three English QA datasets. The comparison controls inputs and instruction detail, not computation, and says nothing about the internal reasoning of the judges. Sonnet-4.6 was run only with C and H, and diagnostics cover smaller slices and mainly two judges. The primary runs used provider defaults for temperature and output length, and the logged model identifiers are aliases whose upstream snapshots we could not resolve (Appendix~\ref{app:reproducibility}). All primary verdicts could be recovered, but many outputs required extracting a JSON object from surrounding text, so this does not show strict format compliance. Sign-test \(p\)-values are not adjusted for multiple comparisons.

Both scope checks provide limited evidence. The 60-row subset has a single annotator, no inter-annotator agreement, and no verified blinding. The generated-answer set has noisy full and partial labels, includes 25 nominal partials identical in content to their full answers, lets \modelopus{} judge its own generations, and used different output caps across judging runs. Neither check estimates how the designs compare against human factuality judgments in general, and extending them remains open.

\section*{Acknowledgments}
Generative AI tools assisted with drafting and editing the text and with checking analysis code and statistical computations. The author is responsible for the content of the paper.

\bibliography{refs}

\clearpage
\appendix

\section{Robustness Details}
\label{app:robustness}

We test three pointwise perturbations on 40 TruthfulQA rows (20 questions) and 50 ASQA rows (25 questions): verbosity padding, a reference distractor, and swapped references. The perturbations are deterministic. \emph{Verbosity padding} appends the fixed sentence ``This answer adds some extra stylistic filler for readability, but it should not change the factual content.'' to the candidate. \emph{Reference distractor} appends the fixed sentence ``Distractor note: some unrelated details may appear in this reference and should not be treated as answer evidence.'' to the reference; it adds no substantive distractor content. \emph{Swap reference} replaces the reference with the reference of another example. Table~\ref{tab:robustness} shows that, among the tested perturbations, swapped references move verdicts the most. TruthfulQA is relatively stable. ASQA is not: both atomic and holistic judges change their verdict on roughly a third to a half of rows when the reference is replaced with one from another example.

\begin{table}[t]
\centering
\small
\setlength{\tabcolsep}{4pt}
\begin{tabular}{llrrrr}
\toprule
Judge & Design & Cons. & Swap & Distr. & Pad \\
\midrule
\multicolumn{6}{l}{\emph{TruthfulQA}} \\
\modelopus{} & C & 0.950 & 0.050 & 0.025 & 0.025 \\
GPT-4.1 & C & 0.975 & 0.000 & 0.025 & 0.025 \\
\modelopus{} & H & 0.975 & 0.025 & 0.000 & 0.000 \\
GPT-4.1 & H & 0.925 & 0.050 & 0.025 & 0.000 \\
\midrule
\multicolumn{6}{l}{\emph{ASQA}} \\
\modelopus{} & C & 0.500 & 0.500 & 0.040 & 0.020 \\
GPT-4.1 & C & 0.520 & 0.480 & 0.040 & 0.060 \\
\modelopus{} & H & 0.500 & 0.500 & 0.080 & 0.040 \\
GPT-4.1 & H & 0.620 & 0.340 & 0.060 & 0.100 \\
\bottomrule
\end{tabular}
\caption{Pointwise robustness on targeted diagnostic slices. \emph{Cons.} is the fraction of rows whose predicted verdict is identical under the identity condition and every perturbation run for that row. \emph{Swap / Distr. / Pad} is the fraction of rows whose verdict under that perturbation differs from the identity verdict. Atomic = candidate-side (C); Holistic = matched holistic (H).}
\label{tab:robustness}
\end{table}

Pairwise order bias is much less severe than reference perturbation in these slices (Table~\ref{tab:order}). Here a pairwise judge compares two candidates, and the candidate-order swap exchanges Answer A and Answer B. TruthfulQA and QAMPARI show no order flips for either judge. On ASQA, GPT-4.1 has zero flips, while \modelopus{} shows a small but non-zero flip rate.

\begin{table}[t]
\centering
\small
\setlength{\tabcolsep}{4pt}
\begin{tabular}{llrr}
\toprule
Dataset & Judge & Order flip & Accuracy \\
\midrule
TruthfulQA & \modelopus{} & 0.00 & 1.00 \\
TruthfulQA & GPT-4.1 & 0.00 & 1.00 \\
ASQA & \modelopus{} & 0.04 & 0.94 \\
ASQA & GPT-4.1 & 0.00 & 1.00 \\
QAMPARI & \modelopus{} & 0.00 & 1.00 \\
QAMPARI & GPT-4.1 & 0.00 & 1.00 \\
\bottomrule
\end{tabular}
\caption{Pairwise order robustness on targeted diagnostic slices (20 question pairs for TruthfulQA; 25 each for ASQA and QAMPARI). \emph{Position Flip Rate} is the fraction of source examples whose selected winning candidate changes when the candidate order is reversed.}
\label{tab:order}
\end{table}

\section{Diagnostic Ablations}
\label{app:diagnostics}

These ablations use small targeted slices and, except for the prompt-variant check, only \modelopus{} and GPT-4.1. Accuracy is row-level agreement with construction labels. Atomic denotes candidate-side decomposition (C) and holistic the matched rubric (H).

\paragraph{Prompt variants.}
Besides the main templates (v1), we wrote two alternative wordings per family, giving three atomic and three holistic variants (exact text in Appendix~\ref{app:prompt-templates}). Each variant keeps the family's design principle (claim decomposition or rubric-based judging) and changes the framing, instruction order, and emphasis. The variants are not symmetric in what they make explicit: atomic v1 does not mention completeness, while atomic v2 does, and all holistic variants address coverage or missing information. Table~\ref{tab:prompt-variants} reports results. On these ASQA slices, every holistic variant is more accurate than every atomic variant for both \modelopus{} and Sonnet-4.6. On TruthfulQA, all six \modelopus{} variants lie between 0.950 and 0.975. The v2 and v3 templates are wordier and cost more tokens.

\begin{table}[t]
\centering
\small
\setlength{\tabcolsep}{4pt}
\begin{tabular}{lllrr}
\toprule
Dataset & Judge & Variant & Acc. & Tokens \\
\midrule
ASQA & \modelopus{} & C v1 & 0.78 & 519.9 \\
ASQA & \modelopus{} & C v2 & 0.78 & 615.9 \\
ASQA & \modelopus{} & C v3 & 0.78 & 643.7 \\
ASQA & \modelopus{} & H v1 & 0.88 & 479.5 \\
ASQA & \modelopus{} & H v2 & 0.94 & 562.9 \\
ASQA & \modelopus{} & H v3 & 0.94 & 568.8 \\
\midrule
ASQA & Sonnet-4.6 & C v2 & 0.76 & 627.1 \\
ASQA & Sonnet-4.6 & C v3 & 0.72 & 646.8 \\
ASQA & Sonnet-4.6 & H v2 & 0.88 & 574.8 \\
ASQA & Sonnet-4.6 & H v3 & 0.88 & 583.9 \\
\midrule
TruthfulQA & \modelopus{} & C v1 & 0.950 & 345.2 \\
TruthfulQA & \modelopus{} & C v2 & 0.975 & 436.8 \\
TruthfulQA & \modelopus{} & C v3 & 0.975 & 464.0 \\
TruthfulQA & \modelopus{} & H v1 & 0.975 & 374.8 \\
TruthfulQA & \modelopus{} & H v2 & 0.975 & 434.7 \\
TruthfulQA & \modelopus{} & H v3 & 0.975 & 452.0 \\
\bottomrule
\end{tabular}
\caption{Prompt-variant check on diagnostic slices: three wordings per method family. Tokens are mean total tokens per call. Sonnet-4.6 was run with v2 and v3 only on this slice.}
\label{tab:prompt-variants}
\end{table}

\paragraph{Schema control.}
The schema-matched judge receives the holistic rubric (``Judge holistically on: correctness, completeness, unsupported detail, robustness to style bias''), is told not to decompose the answer into atomic claims, and must return the candidate-side atomic key names, filled with short evidence notes. It asks whether atomic results, where they differ, come mainly from the output format. On TruthfulQA, its accuracy matches C in the separate formal diagnostic runs: .950 for \modelopus{} and .975 for GPT-4.1, at higher token cost for both. On ASQA, its .800/.620 lies between C's .780/.580 and H's .880/.720 in those same formal diagnostic runs. These comparators use the same rows as the schema control, but they are separate from the budget-ablation reruns; the 256-token budget run, for example, gives GPT-4.1 C an accuracy of 1.000 rather than .975 on TruthfulQA.

\begin{table}[t]
\centering
\small
\setlength{\tabcolsep}{4pt}
\begin{tabular}{llrr}
\toprule
Dataset & Judge & Acc. & Total tokens \\
\midrule
TruthfulQA & \modelopus{} & 0.950 & 424.85 \\
TruthfulQA & GPT-4.1 & 0.975 & 345.78 \\
ASQA & \modelopus{} & 0.800 & 567.48 \\
ASQA & GPT-4.1 & 0.620 & 477.68 \\
\bottomrule
\end{tabular}
\caption{Schema control: atomic key names without the claim-decomposition instruction. Targeted TruthfulQA and ASQA diagnostic slices (40 TruthfulQA rows (20 questions) and 50 ASQA rows (25 questions)).}
\label{tab:schema-control}
\end{table}

\paragraph{Budget control.}
The budget ablation sets an explicit output-token cap of 128, 256, or 512 and asks whether atomic judging simply needs more room to write. On ASQA, raising the cap does not close the gap for either judge; on TruthfulQA, accuracy stays nearly flat across caps (Table~\ref{tab:budget-control}).

\begin{table}[t]
\centering
\small
\setlength{\tabcolsep}{4pt}
\begin{tabular}{lllrr}
\toprule
Dataset & Design & Cap & \modelopus{} & GPT-4.1 \\
\midrule
TruthfulQA & C & 128 & 0.975 & 0.975 \\
TruthfulQA & C & 256 & 0.950 & 1.000 \\
TruthfulQA & C & 512 & 0.950 & 0.975 \\
TruthfulQA & H & 128 & 0.975 & 0.975 \\
TruthfulQA & H & 256 & 0.975 & 0.975 \\
TruthfulQA & H & 512 & 0.975 & 0.975 \\
ASQA & C & 128 & 0.760 & 0.580 \\
ASQA & C & 256 & 0.760 & 0.600 \\
ASQA & C & 512 & 0.760 & 0.600 \\
ASQA & H & 128 & 0.880 & 0.660 \\
ASQA & H & 256 & 0.880 & 0.680 \\
ASQA & H & 512 & 0.860 & 0.700 \\
\bottomrule
\end{tabular}
\caption{Budget control: Cap is the requested maximum output tokens per call. Cells give row-level accuracy under explicit output caps of 128, 256, and 512 tokens on targeted TruthfulQA and ASQA diagnostic slices (40 TruthfulQA rows (20 questions) and 50 ASQA rows (25 questions)). Raising the cap does not reverse the ASQA ranking.}
\label{tab:budget-control}
\end{table}

\paragraph{Grounding control.}
The grounding ablation asks how much the judges rely on the supplied reference. \emph{Reduced reference} is a deterministic delimiter-based truncation, not a paraphrase. The constructor tries the separators ``. '', ``? '', and ``! '' in that order and truncates at the first occurrence of the first separator it finds; with none present, it leaves the text unchanged. \emph{No reference} removes the reference, and the renderer then fills the reference slot with the literal string ``(missing)''. Each condition has 40 rows from 20 questions on TruthfulQA and 50 rows from 25 questions on ASQA. TruthfulQA accuracy is unchanged under the reduced reference and drops once the reference is removed. ASQA is much more sensitive: even the reduced reference lowers accuracy by 20--24 points (Table~\ref{tab:grounding-control}). Construction labels are held fixed when references change. The scores therefore measure agreement with the original labels, not correctness relative to the altered evidence. We read this as reference sensitivity, not as a causal account of the C--H gap.

\begin{table}[t]
\centering
\small
\setlength{\tabcolsep}{4pt}
\begin{tabular}{llrrr}
\toprule
Judge & Design & Full & Reduced & None \\
\midrule
\multicolumn{5}{l}{\emph{TruthfulQA}} \\
\modelopus{} & C & 0.950 & 0.950 & 0.625 \\
GPT-4.1 & C & 0.975 & 0.975 & 0.750 \\
\modelopus{} & H & 0.975 & 0.975 & 0.550 \\
GPT-4.1 & H & 0.975 & 0.975 & 0.700 \\
\midrule
\multicolumn{5}{l}{\emph{ASQA}} \\
\modelopus{} & C & 0.740 & 0.500 & 0.380 \\
GPT-4.1 & C & 0.540 & 0.340 & 0.260 \\
\modelopus{} & H & 0.860 & 0.620 & 0.380 \\
GPT-4.1 & H & 0.700 & 0.500 & 0.480 \\
\bottomrule
\end{tabular}
\caption{Grounding control: row-level accuracy with the full reference (identity), the delimiter-truncated reference, and no reference. 40 rows (20 questions) per condition on TruthfulQA; 50 rows (25 questions) on ASQA.}
\label{tab:grounding-control}
\end{table}

\section{Generated-Answer ASQA Set}
\label{app:llmgen}

\paragraph{Construction.}
The set covers 50 ASQA source questions, each answered by two generators, \modelopus{} and Sonnet-4.6, for 100 generated answers. The generation code defaults to temperature 0.7 and a 256-token output cap, but the generated-answer records do not retain per-call decoding settings. Each generated answer gives two rows: the full answer, labeled \texttt{supported}, and a nominal partial answer, labeled \texttt{partially\_supported}. That makes 200 rows, 100 per generator. The partial constructor splits the answer literally on ``. '', keeps the first third of the segments, and appends a period; it is not a sentence-aware truncator. For 25 of the 100 answers (10 \modelopus{}, 15 Sonnet-4.6), the nominal partial string is the full answer plus an appended period. Both construction labels are therefore noisy. The \texttt{supported} label assumes that a free-form answer covers the ASQA reference, and many partial strings are not truncated at all. These results are not a reliably labeled test of free-form factuality accuracy.

\paragraph{Judging settings.}
In the generated-set judging logs, \modelopus{} C and H calls carry an explicit 256-token output cap, whereas \modelopus{} R and all GPT-4.1 and Gemini Flash Lite C/H/R calls omit the cap. No call in any of the nine judge--design combinations sets a temperature. Settings are therefore not uniform across the generated-set comparisons.

\paragraph{Results.}
Table~\ref{tab:llmgen} pools both generators, and Table~\ref{tab:llmgen-split} splits the results by generator. Pooled, H is more accurate than C for all three judges (\modelopus{} 95 vs.\ 88 of 200 rows correct; GPT-4.1 113 vs.\ 76; Gemini Flash Lite 123 vs.\ 103), and H is also ahead of C within each generator for every judge. R has the highest nominal-partial accuracy (74--78\%) but the lowest full-answer accuracy (16--38\%): it often labels complete-looking generated answers \texttt{partially\_supported} against the multi-aspect ASQA reference. \modelopus{} also judges its own 50 generations, which appear as 100 rows. The generator split is small and confounded with this overlap and with the label noise above, so we draw no conclusion about the presence or absence of self-preference.

\begin{table}[t]
\centering
\small
\begin{tabular}{llrrr}
\toprule
Judge & Design & Full & Partial & Overall \\
\midrule
\modelopus{} & C & 34.0 & 54.0 & 44.0 \\
 & H & 30.0 & 65.0 & 47.5 \\
 & R & 19.0 & 78.0 & 48.5 \\
\midrule
GPT-4.1 & C & 48.0 & 28.0 & 38.0 \\
 & H & 64.0 & 49.0 & 56.5 \\
 & R & 23.0 & 75.0 & 49.0 \\
\midrule
Gemini FL & C & 66.0 & 37.0 & 51.5 \\
 & H & 74.0 & 49.0 & 61.5 \\
 & R & 36.0 & 76.0 & 56.0 \\
\bottomrule
\end{tabular}
\caption{Generated-answer ASQA set, both generators pooled: accuracy (\%) on the 100 full rows, the 100 nominal-partial rows, and all 200 rows. 50 source questions $\times$ 2 generators = 100 answers = 200 rows. Both construction labels are noisy; 25 of the 100 nominal partials equal the full answer plus a period.}
\label{tab:llmgen}
\end{table}

\begin{table*}[t]
\centering
\small
\setlength{\tabcolsep}{4pt}
\begin{tabular}{llccc}
\toprule
Judge & Generator & C & H & R \\
\midrule
\modelopus{} & \modelopus{} & 34 / 54 / 44 & 32 / 66 / 49 & 22 / 78 / 50 \\
\modelopus{} & Sonnet-4.6 & 34 / 54 / 44 & 28 / 64 / 46 & 16 / 78 / 47 \\
GPT-4.1 & \modelopus{} & 44 / 24 / 34 & 66 / 52 / 59 & 26 / 76 / 51 \\
GPT-4.1 & Sonnet-4.6 & 52 / 32 / 42 & 62 / 46 / 54 & 20 / 74 / 47 \\
Gemini FL & \modelopus{} & 68 / 34 / 51 & 78 / 48 / 63 & 38 / 74 / 56 \\
Gemini FL & Sonnet-4.6 & 64 / 40 / 52 & 70 / 50 / 60 & 34 / 78 / 56 \\
\bottomrule
\end{tabular}
\caption{Generated-answer ASQA set by generator. Cells give full / nominal-partial / overall accuracy (\%) over 50 / 50 / 100 rows per generator. The \modelopus{}-judge, \modelopus{}-generator rows are the judge's own 50 generations.}
\label{tab:llmgen-split}
\end{table*}

\section{Author Label Check}
\label{app:human-annotations}

\paragraph{Design.}
One author labeled 60 rows: ten two-candidate pairs (20 rows) from each of the three datasets, 30 source questions in total. Each row received one of \texttt{supported}, \texttt{partially\_supported}, or \texttt{unsupported}, judged as factual correctness of the candidate given the question and reference. This is a single-annotator check. There is no inter-annotator agreement, and we do not claim that the annotation was blinded to construction labels or judge outputs.

\paragraph{Two separate comparisons.}
Table~\ref{tab:human-sanity} reports two things that should not be conflated. The upper panel compares construction labels with the author's labels. They agree on 20/20 TruthfulQA and 20/20 QAMPARI rows and on 11/20 ASQA rows (51/60 overall). All nine disagreements are ASQA rows that the construction labels \texttt{partially\_supported} and the author labels \texttt{supported}; Table~\ref{tab:human-detail} lists them. The lower panel scores each judge against each standard. Under construction labels, H is more accurate than C for all three judges. Under the author's labels, C is more accurate than H for all three judges. R is the most accurate design under construction labels (59, 56, 55) and the least accurate under the author's labels (50, 51, 52). These are descriptive differences of one to nine rows out of 60. They are not population-level superiority tests or evidence that either label set is reliable.

\begin{table}[t]
\centering
\small
\setlength{\tabcolsep}{4pt}
\begin{tabular}{lccc}
\toprule
\multicolumn{4}{l}{\emph{Construction vs.\ author labels (agreement)}} \\
\midrule
Dataset & Rows & Agree & \% \\
\midrule
TruthfulQA & 20 & 20 & 100.0 \\
ASQA & 20 & 11 & 55.0 \\
QAMPARI & 20 & 20 & 100.0 \\
Overall & 60 & 51 & 85.0 \\
\midrule
\multicolumn{4}{l}{\emph{Judge rows correct (of 60): construction / author}} \\
\midrule
Judge & C & H & R \\
\midrule
\modelopus{} & 53 / 54 & 55 / 52 & 59 / 50 \\
GPT-4.1 & 49 / 54 & 50 / 53 & 56 / 51 \\
Gemini FL & 48 / 55 & 50 / 53 & 55 / 52 \\
\bottomrule
\end{tabular}
\caption{Single-author label check on 60 rows (30 source questions; ten two-candidate pairs per dataset). Upper: agreement between construction and author labels; all nine disagreements are ASQA construction-\texttt{partially\_supported} rows labeled \texttt{supported} by the author. Lower: judge rows correct out of 60 against construction labels and against the author's labels.}
\label{tab:human-sanity}
\end{table}

\begin{table*}[t]
\centering
\small
\resizebox{\textwidth}{!}{
\begin{tabular}{p{4.2cm} p{6.4cm} p{4.6cm}}
\toprule
Question & Candidate (exact; ``\ldots'' marks an omission) & Reference content the candidate omits \\
\midrule
When was the 13th amendment ratified by the states? & ``When was the 13th amendment ratified by the required number of states? December 6, 1865'' & Passage on January 31, 1865; 18 states by end of February 1865 \\
Who played zordon in the original power rangers? & ``Who played Zordon in the earlier episodes of the original power rangers TV series? David Fielding'' & Later voice actor; 1995 film portrayal \\
When is episode 113 of dragon ball super coming out? & ``When is episode 113 of dragon ball super coming out for its original airdate? October 29, 2017 \ldots{} for its american airdate? June 1, 2019'' & Series start date (July 5, 2015) \\
Which political party is currently in power in ireland? & ``\ldots{} from 2011 to 2016? Fine Gael--Labour Party coalition \ldots{} from 2016 to 2020? Fine Gael \ldots{} presidency from 2011 to 2018? Labour Party'' & Earlier governments and coalition history from 1997 \\
Japanese hotel run by same family for 1300 years? & ``Japanese hotel run by same family for 1300 years in Komatsu? H\=oshi Ryokan'' & Nishiyama Onsen Keiunkan \\
Who killed the man in thelma and louise? & ``Which character killed the man in thelma and louise? Louise Elizabeth Sawyer Which actor killed the man in thelma and louise? Susan Sarandon'' & Narrative detail of the shooting \\
The creation of the senate was based on the? & ``The creation of the US senate was based on which ancient body? Roman Senate'' & Other bases covered by the reference \\
When is fortnite battle royale being released on android? & ``When is fortnite battle royale beta version being released on android for samsung devices exclusively? August 9, 2018'' & Non-Samsung invites (August 13, 2018) and general availability (October 11, 2018) \\
Present chairman of public accounts committee of lok sabha? & ``In 2017, who was the chairman of public accounts committee of lok sabha? Mallikarjun Kharge'' & Chairs in 2015 and 2016 \\
\bottomrule
\end{tabular}
}
\caption{The nine ASQA disagreement rows. In each case the construction label is \texttt{partially\_supported} because the candidate covers only part of the reference, while the author judged the answer \texttt{supported}. Candidates follow the ASQA partial construction (``sub-question + first short answer''). The right column summarizes omitted reference content; it is not the full reference text.}
\label{tab:human-detail}
\end{table*}

\section{Additional Statistics}
\label{app:stats}

\paragraph{Class-wise accuracy.}
Table~\ref{tab:class} reports accuracy on the discriminating class: unsupported rows for TruthfulQA and partially supported rows for ASQA and QAMPARI. Each class contains 200 rows. Supported-row accuracy ranges from .940 to 1.000 across all evaluated combinations. On ASQA, the H--C partial-row differences are +14.5, +30.5, +33.0, and +19.0 points for \modelopus{}, GPT-4.1, Gemini Flash Lite, and Sonnet-4.6, respectively. On TruthfulQA, most of the candidate-side advantage comes from unsupported rows.

\begin{table}[t]
\centering
\small
\setlength{\tabcolsep}{5pt}
\begin{tabular}{lrrrr}
\toprule
Judge & C & H & R & B \\
\midrule
\multicolumn{5}{l}{\emph{TruthfulQA: unsupported}} \\
\modelopus{} & .950 & .940 & .970 & .960 \\
GPT-4.1 & .960 & .940 & .955 & .960 \\
Gemini FL & .930 & .870 & .915 & .935 \\
Sonnet-4.6 & .950 & .890 & --- & --- \\
\midrule
\multicolumn{5}{l}{\emph{ASQA: partially supported}} \\
\modelopus{} & .555 & .700 & .950 & .800 \\
GPT-4.1 & .160 & .465 & .870 & .610 \\
Gemini FL & .125 & .455 & .745 & .520 \\
Sonnet-4.6 & .430 & .620 & --- & --- \\
\midrule
\multicolumn{5}{l}{\emph{QAMPARI: partially supported}} \\
\modelopus{} & .950 & .985 & .990 & .965 \\
GPT-4.1 & .865 & .995 & 1.000 & .995 \\
Gemini FL & .905 & .995 & .995 & .990 \\
Sonnet-4.6 & .925 & .990 & --- & --- \\
\bottomrule
\end{tabular}
\caption{Accuracy on each dataset's discriminating class (200 rows per cell). Sonnet-4.6 was evaluated with C and H only.}
\label{tab:class}
\end{table}

\paragraph{Output-token distributions.}
Table~\ref{tab:lengths} gives realized output tokens for C and H in the primary runs for \modelopus{}, GPT-4.1, and Sonnet-4.6. The Gemini interface recorded total tokens but not output tokens. Candidate-side outputs are longer than holistic outputs on ASQA and especially on QAMPARI, where they enumerate long answer lists.

\begin{table*}[t]
\centering
\small
\begin{tabular}{llcc}
\toprule
Dataset & Judge & C: median [Q1, Q3] & H: median [Q1, Q3] \\
\midrule
TruthfulQA & \modelopus{} & 141.00 [119.00, 165.00] & 137.50 [103.75, 198.25] \\
TruthfulQA & GPT-4.1 & 106.00 [99.00, 116.00] & 104.00 [88.75, 127.00] \\
TruthfulQA & Sonnet-4.6 & 153.50 [132.00, 187.00] & 173.00 [115.00, 217.25] \\
\midrule
ASQA & \modelopus{} & 249.50 [207.00, 307.00] & 149.50 [110.00, 219.00] \\
ASQA & GPT-4.1 & 182.00 [144.00, 233.00] & 113.00 [104.00, 138.00] \\
ASQA & Sonnet-4.6 & 272.00 [224.00, 318.00] & 161.50 [132.00, 226.00] \\
\midrule
QAMPARI & \modelopus{} & 339.00 [264.00, 501.00] & 156.50 [97.75, 201.25] \\
QAMPARI & GPT-4.1 & 268.50 [196.00, 403.25] & 117.50 [95.00, 148.00] \\
QAMPARI & Sonnet-4.6 & 348.00 [274.00, 505.00] & 160.50 [118.00, 209.00] \\
\bottomrule
\end{tabular}
\caption{Output tokens per call, with median and quartiles from 400 observations per cell. Quantiles use linear interpolation (Hyndman--Fan type 7). Gemini is omitted because output-token breakdowns were not recorded, not because verdicts or total tokens are missing.}
\label{tab:lengths}
\end{table*}

\paragraph{Reference-side versus holistic on ASQA.}
Table~\ref{tab:ref-sign} gives the exact two-sided source-level sign test for R against H on ASQA, computed like the C--H tests in the main text: for each of the 200 questions, count the rows each design labels correctly, and test how the questions on which they differ split between them. Token differences use the mean total tokens of the main results table.

\begin{table}[t]
\centering
\small
\setlength{\tabcolsep}{4pt}
\begin{tabular}{lrrrr}
\toprule
Judge & $\Delta$Acc. & $\Delta$Tokens & R:H & $p$ \\
\midrule
\modelopus{} & +0.1250 & +138.39 & 50:0 & 1.78e-15 \\
GPT-4.1 & +0.2125 & +116.26 & 84:1 & 4.45e-24 \\
Gemini FL & +0.1700 & +123.23 & 69:2 & 2.17e-18 \\
\bottomrule
\end{tabular}
\caption{ASQA reference-side (R) minus holistic (H). R:H counts questions (of 200) on which R or H labels more rows correctly; $p$ is the exact two-sided sign test on those questions, unadjusted for multiple comparisons. Token differences are computed before rounding and are 28--29\% of the holistic mean.}
\label{tab:ref-sign}
\end{table}

\paragraph{Source-clustered bootstrap.}
Table~\ref{tab:bootstrap} reports source-clustered uncertainty summaries computed from the stored judge outputs. Rows are grouped by source question, and each question contributes the mean correctness and mean total tokens of its two rows. Source-question keys are sorted lexicographically before resampling. Differences are H minus C. The bootstrap draws 10{,}000 resamples of 200 source questions with replacement, sharing indices between the accuracy and token differences, and reports percentile intervals (linear interpolation, Hyndman--Fan type 7). Sign stability draws 1{,}000 subsets of 100 of the 200 questions without replacement and counts the fraction whose mean difference is strictly nonzero and has the same sign as the full-data difference. Draws with zero difference do not count as preserving the sign (233, 48, 0, 0, 12, and 0 draws in table order). The random generator is NumPy PCG64 with seed 42, reset for each dataset--judge pair and separately for the bootstrap and the stability draws. The percentile interval and the exact sign test are different procedures. For TruthfulQA GPT-4.1 the interval excludes zero while the sign test is not significant ($p=.125$), and we do not treat the two as equivalent significance tests.

\begin{table*}[t]
\centering
\small
\begin{tabular}{llrrrrr}
\toprule
Dataset & Judge & $\Delta$Acc. & 95\% CI & $\Delta$Tok. & 95\% CI & Stab. \\
\midrule
TruthfulQA & \modelopus{} & $-$.0050 & [$-$.0150, +.0050] & +28.27 & [+25.93, +30.63] & .694 \\
TruthfulQA & GPT-4.1 & $-$.0100 & [$-$.0200, $-$.0025] & +7.92 & [+6.25, +9.52] & .952 \\
ASQA & \modelopus{} & +.0725 & [+.0475, +.0975] & $-$77.11 & [$-$84.29, $-$70.52] & 1.000 \\
ASQA & GPT-4.1 & +.1550 & [+.1200, +.1925] & $-$65.07 & [$-$70.48, $-$59.84] & 1.000 \\
QAMPARI & \modelopus{} & +.0175 & [+.0050, +.0300] & $-$292.79 & [$-$343.09, $-$247.36] & .988 \\
QAMPARI & GPT-4.1 & +.0700 & [+.0475, +.0950] & $-$256.39 & [$-$304.02, $-$214.09] & 1.000 \\
\bottomrule
\end{tabular}
\caption{Source-clustered bootstrap for holistic (H) minus candidate-side (C): 200 source questions per cell, 10{,}000 resamples, 95\% percentile intervals. Accuracy differences are proportions. \emph{Stab.}\ is the fraction of 1{,}000 half-sized subsets (100 questions, without replacement) whose mean difference is strictly nonzero and has the full-data sign.}
\label{tab:bootstrap}
\end{table*}

\paragraph{Complexity bins.}
Table~\ref{tab:bins} stratifies the H minus C accuracy difference by a reference-complexity proxy: the number of ASQA \texttt{qa\_pairs} and the number of QAMPARI gold answers. These counts are proxies, not measured claim inventories. Each question contributes two rows. Across the 24 bin--judge cells for the three primary judges, 21 favor H, three are ties, and none favors C. ASQA shows no monotone trend. On QAMPARI, the GPT-4.1 and Gemini Flash Lite differences are larger for questions with eight or more answers, up to +12.3 points. This is descriptive heterogeneity within the inspected bins, not evidence that every complexity regime favors H.

\begin{table}[t]
\centering
\small
\setlength{\tabcolsep}{4pt}
\begin{tabular}{lrrrr}
\toprule
\multicolumn{5}{l}{\emph{ASQA}, \texttt{num\_qa\_pairs} (rows)} \\
Judge & 2 (144) & 3 (86) & 4 (94) & 5+ (76) \\
\midrule
\modelopus{} & +.076 & +.058 & +.106 & +.039 \\
GPT-4.1 & +.174 & +.233 & +.032 & +.184 \\
Gemini FL & +.181 & +.221 & +.074 & +.092 \\
\midrule
\multicolumn{5}{l}{\emph{QAMPARI}, \texttt{num\_answers} (rows)} \\
Judge & 5 (90) & 6--7 (92) & 8--9 (56) & 10+ (162) \\
\midrule
\modelopus{} & +.011 & .000 & +.036 & +.025 \\
GPT-4.1 & .000 & +.022 & +.107 & +.123 \\
Gemini FL & +.011 & .000 & +.107 & +.056 \\
\bottomrule
\end{tabular}
\caption{Holistic (H) minus candidate-side (C) accuracy by complexity bin for the three primary judges; row counts per bin in parentheses. 21 of 24 cells favor H and 3 are ties.}
\label{tab:bins}
\end{table}

\paragraph{Output parsing.}
\label{app:parse}
The parser first tries to read the whole output as a JSON object, then extracts an embedded brace-delimited JSON object, and finally falls back to regular-expression recovery of the verdict. Table~\ref{tab:parse} reports, for the three primary judge families, which branch recovered each verdict. Every stored verdict is one of the three canonical labels after parsing. Only 5{,}686 of the 7{,}200 primary C/H outputs, however, are whole-output JSON objects. Canonical verdicts therefore show that labels were recoverable, not that outputs complied with strict JSON.

\begin{table*}[t]
\centering
\small
\begin{tabular}{lrrrrr}
\toprule
Design & Rows & Whole JSON & Embedded JSON & Regex only & Canonical verdict \\
\midrule
C & 3600 & 2765 & 833 & 2 & 3600 \\
H & 3600 & 2921 & 679 & 0 & 3600 \\
R & 3600 & 2492 & 1106 & 2 & 3600 \\
B & 3600 & 2414 & 1181 & 5 & 3600 \\
\midrule
C + H & 7200 & 5686 & 1512 & 2 & 7200 \\
\bottomrule
\end{tabular}
\caption{Parser branch that recovered each verdict, for \modelopus{}, GPT-4.1, and Gemini Flash Lite over the three datasets (1{,}200 rows per judge and design). Sonnet-4.6 rows are not included in these counts.}
\label{tab:parse}
\end{table*}

\section{Qualitative Examples}
\label{app:qualitative}

The following cases are selected from the primary \modelopus{} runs. All question, reference, and candidate texts are reproduced in full, and verdicts come from the saved outputs. They illustrate disagreements over the construction labels, not independently annotated factual errors.

\paragraph{ASQA.}
\textbf{Question:} When did construction start on the sydney harbour bridge?

\textbf{Reference:} Construction of the sydney harbour bridge started on 28 July 1923 but construction of the Arch did not begin until 26 October 1928.

\textbf{Candidate:} When did construction officially start on the sydney harbour bridge? 28 July 1923

\textbf{Construction label:} \texttt{partially\_supported}.

\textbf{Recorded verdicts:} C: supported, H: partially supported, R: partially supported.

The ASQA candidate gives the official construction start date but omits the arch's start date. H's complete parsed output follows (whitespace formatted for display). Accuracy uses only its verdict, not the score, confidence, or rationale.

\begin{lstlisting}[basicstyle=\ttfamily\footnotesize,breaklines=true,columns=fullflexible,keepspaces=true,showstringspaces=false]
{
  "verdict": "partially_supported",
  "score": 4,
  "evidence_summary": "The candidate correctly states that construction started on 28 July 1923, which matches the reference answer.",
  "unsupported_or_missing_points": [
    "Missing detail that construction of the Arch did not begin until 26 October 1928"
  ],
  "confidence": 0.95,
  "rationale": "The core date of 28 July 1923 is correct and directly answers the question. The candidate omits the additional detail about the Arch construction starting on 26 October 1928, which provides useful context but is supplementary information. The answer is correct but not fully complete."
}
\end{lstlisting}

\paragraph{TruthfulQA.}
\textbf{Question:} What side of the road do you have to drive on?

\textbf{Reference:} It depends on which country you are in

\textbf{Candidate:} You have to drive on the right side of the road

\textbf{Construction label:} \texttt{unsupported}.

\textbf{Recorded verdicts:} C: unsupported, H: partially supported, R: unsupported.

\paragraph{QAMPARI.}
\textbf{Question:} What was something that Saipa manufactured?

\textbf{Reference:} Saipa Shahin, Saipa Tiba, Saipa Saina, Saipa Quick, Saipa 151

\textbf{Candidate:} Saipa Shahin, Saipa Tiba

\textbf{Construction label:} \texttt{partially\_supported}.

\textbf{Recorded verdicts:} C: supported, H: partially supported, R: partially supported.

The TruthfulQA reference qualifies the answer by country, whereas the candidate makes an unconditional statement. The QAMPARI candidate lists two supported items but omits three reference items. Source identifiers are ASQA question 2771414991157063420, TruthfulQA question 116, and QAMPARI question \texttt{919\_\_wikidata\_simple\_\_dev}.

\section{Judge Prompt Templates}
\label{app:prompt-templates}
\label{app:prompts}

This appendix reproduces the system prompts and user-message templates as they appear in the project's prompt module (\texttt{src/llm\_as\_judge\_lab/prompts.py}). Each call sends one system message and one user message. The renderer fills \{question\}, \{reference\_answer\}, and \{candidate\_answer\}; when no reference is supplied (grounding control) it substitutes the literal string ``(missing)''. The renderer also supports an optional strictness prefix, but the primary run records carry the \texttt{default} variant, which adds no prefix. The saved prompts of the 16{,}800 core and Sonnet-4.6 rows match these rendered templates and the run manifests (Appendix~\ref{app:reproducibility}). We make no claim about the exact strings used in every earlier development rerun.

\paragraph{Output fields.}
Every pointwise template requests a JSON object whose first two keys are \texttt{verdict} (one of the three labels) and \texttt{score} (integer 1--5) and whose last two are \texttt{confidence} and \texttt{rationale}. Accuracy uses only the parsed \texttt{verdict}. Score, confidence, rationale, and claim lists do not enter accuracy. The \texttt{score} field is used only for score-shift summaries in some robustness diagnostics.

\subsection{Shared user-message header}
Every pointwise user message begins with the following lines. The design-specific block follows after a blank line.

\begin{quote}
\small\ttfamily\raggedright
Question:\\
\{question\}\\[3pt]
Reference answer:\\
\{reference\_answer\}\\[3pt]
Candidate answer:\\
\{candidate\_answer\}
\end{quote}

\subsection{C: candidate-side atomic (v1, main)}
\label{app:atomic-v1}
\paragraph{System prompt.}
\begin{quote}\small
You are a careful factuality judge. Break the candidate answer into atomic factual claims, ground each claim in the reference answer or context, and then produce a structured verdict. Ignore verbosity, tone, or writing style unless they change factual content.
\end{quote}
\paragraph{User block.}
\begin{quote}
\small\ttfamily\raggedright
Return valid JSON with keys:\\
- verdict: one of ["supported", "partially\_supported", "unsupported"]\\
- score: integer 1-5\\
- atomic\_claims: list of strings\\
- supported\_claims: list of strings\\
- unsupported\_claims: list of strings\\
- confidence: number between 0 and 1\\
- rationale: short string
\end{quote}

\subsection{H: matched holistic (v1, main)}
\label{app:holistic-v1}
\paragraph{System prompt.}
\begin{quote}\small
You are a careful factuality judge. Evaluate the candidate answer against the reference answer or context with a detailed rubric, but do not explicitly decompose the answer into atomic claims. Ignore style or verbosity unless they change factual content.
\end{quote}
\paragraph{User block.}
\begin{quote}
\small\ttfamily\raggedright
Judge holistically on:\\
- correctness\\
- completeness\\
- unsupported detail\\
- robustness to style bias\\[3pt]
Return valid JSON with keys:\\
- verdict: one of ["supported", "partially\_supported", "unsupported"]\\
- score: integer 1-5\\
- evidence\_summary: short string\\
- unsupported\_or\_missing\_points: list of strings\\
- confidence: number between 0 and 1\\
- rationale: short string
\end{quote}

\subsection{R: reference-side atomic (main)}
\paragraph{System prompt.}
\begin{quote}\small
You are a careful factuality judge. Break the reference answer into atomic factual claims, check for each claim whether it is covered by the candidate answer, and then produce a structured verdict. Ignore verbosity, tone, or writing style unless they change factual content.
\end{quote}
\paragraph{User block.}
\begin{quote}
\small\ttfamily\raggedright
Return valid JSON with keys:\\
- verdict: one of ["supported", "partially\_supported", "unsupported"]\\
- score: integer 1-5\\
- reference\_claims: list of strings\\
- covered\_claims: list of strings\\
- missing\_claims: list of strings\\
- confidence: number between 0 and 1\\
- rationale: short string
\end{quote}

\subsection{B: bidirectional atomic (main)}
\paragraph{System prompt.}
\begin{quote}\small
You are a careful factuality judge. Break both the candidate answer and the reference answer into atomic factual claims. Ground each candidate claim in the reference answer, and check for each reference claim whether it is covered by the candidate answer. Then produce a structured verdict. Ignore verbosity, tone, or writing style unless they change factual content.
\end{quote}
\paragraph{User block.}
\begin{quote}
\small\ttfamily\raggedright
Return valid JSON with keys:\\
- verdict: one of ["supported", "partially\_supported", "unsupported"]\\
- score: integer 1-5\\
- candidate\_claims: list of strings\\
- supported\_claims: list of strings\\
- unsupported\_claims: list of strings\\
- reference\_claims: list of strings\\
- covered\_claims: list of strings\\
- missing\_claims: list of strings\\
- confidence: number between 0 and 1\\
- rationale: short string
\end{quote}

\subsection{Candidate-side atomic v2}
\paragraph{System prompt.}
\begin{quote}\small
You are a meticulous fact-checking assistant. Your task is to identify every key factual assertion in the candidate answer, verify each assertion against the reference answer, and then aggregate your findings into a structured JSON verdict. Focus exclusively on factual accuracy and completeness; disregard differences in phrasing, tone, or length.
\end{quote}
\paragraph{User block.}
\begin{quote}
\small\ttfamily\raggedright
Perform the following analysis:\\
1. Identify every key factual assertion in the candidate answer.\\
2. For each assertion, determine whether it is supported by the reference answer.\\
3. Aggregate your findings into an overall verdict.\\[3pt]
Return valid JSON with keys:\\
- verdict: one of ["supported", "partially\_supported", "unsupported"]\\
- score: integer 1-5\\
- atomic\_claims: list of strings (all factual assertions identified)\\
- supported\_claims: list of strings (assertions backed by the reference)\\
- unsupported\_claims: list of strings (assertions not backed or contradicted)\\
- confidence: number between 0 and 1\\
- rationale: short string explaining the overall judgement
\end{quote}

\subsection{Candidate-side atomic v3}
\paragraph{System prompt.}
\begin{quote}\small
You are a rigorous claim-verification judge. Follow a strict step-by-step protocol: first list all discrete claims made in the candidate answer, then check each claim for support in the reference answer, and finally render an overall verdict with a confidence score. Ignore stylistic differences and evaluate only factual substance.
\end{quote}
\paragraph{User block.}
\begin{quote}
\small\ttfamily\raggedright
Follow these steps strictly:\\
Step 1: List every discrete claim made in the candidate answer.\\
Step 2: For each claim, check whether the reference answer supports, contradicts, or is silent on it.\\
Step 3: Based on your verification, decide on an overall verdict and score.\\[3pt]
Return valid JSON with keys:\\
- verdict: one of ["supported", "partially\_supported", "unsupported"]\\
- score: integer 1-5\\
- atomic\_claims: list of strings (all discrete claims from Step 1)\\
- supported\_claims: list of strings (claims supported in Step 2)\\
- unsupported\_claims: list of strings (claims contradicted or unsupported in Step 2)\\
- confidence: number between 0 and 1\\
- rationale: short string summarizing your Step 3 reasoning
\end{quote}

\subsection{Holistic v2}
\paragraph{System prompt.}
\begin{quote}\small
You are an expert answer-quality evaluator. Assess the candidate answer by first analyzing how well it covers the information in the reference answer, then checking for factual accuracy, and finally assigning an overall quality score. Do not decompose the answer into atomic claims. Disregard style and verbosity unless they introduce factual errors.
\end{quote}
\paragraph{User block.}
\begin{quote}
\small\ttfamily\raggedright
Evaluate the candidate answer holistically in three stages:\\
1. Coverage: How much of the reference information does the candidate include?\\
2. Accuracy: Are the included facts correct relative to the reference?\\
3. Final score: Combine coverage and accuracy into an overall assessment.\\[3pt]
Return valid JSON with keys:\\
- verdict: one of ["supported", "partially\_supported", "unsupported"]\\
- score: integer 1-5\\
- evidence\_summary: short string describing coverage and accuracy findings\\
- unsupported\_or\_missing\_points: list of strings (missing or incorrect information)\\
- confidence: number between 0 and 1\\
- rationale: short string explaining the final score
\end{quote}

\subsection{Holistic v3}
\paragraph{System prompt.}
\begin{quote}\small
You are a comparative factuality judge. Your job is to compare the candidate answer directly against the reference answer, evaluating factual overlap, missing information, and any unsupported additions. Produce a single holistic judgement without breaking the answer into atomic claims. Ignore differences in writing style or length.
\end{quote}
\paragraph{User block.}
\begin{quote}
\small\ttfamily\raggedright
Compare the candidate answer against the reference answer on these dimensions:\\
- Factual overlap: What correct information do both share?\\
- Missing information: What does the reference contain that the candidate omits?\\
- Unsupported additions: What does the candidate assert that the reference does not support?\\[3pt]
Synthesize your comparison into an overall verdict.\\[3pt]
Return valid JSON with keys:\\
- verdict: one of ["supported", "partially\_supported", "unsupported"]\\
- score: integer 1-5\\
- evidence\_summary: short string summarizing the comparison\\
- unsupported\_or\_missing\_points: list of strings (omissions and unsupported additions)\\
- confidence: number between 0 and 1\\
- rationale: short string explaining the verdict
\end{quote}

\subsection{Schema-matched control}
\paragraph{System prompt.}
\begin{quote}\small
You are a careful factuality judge. Evaluate the candidate answer against the reference answer or context, but do not explicitly decompose the answer into atomic claims. Use the requested JSON fields as compact evidence notes rather than as an exhaustive claim list. Ignore verbosity, tone, or writing style unless they change factual content.
\end{quote}
\paragraph{User block.}
\begin{quote}
\small\ttfamily\raggedright
Judge holistically on:\\
- correctness\\
- completeness\\
- unsupported detail\\
- robustness to style bias\\[3pt]
Return valid JSON with the same keys as the atomic judge:\\
- verdict: one of ["supported", "partially\_supported", "unsupported"]\\
- score: integer 1-5\\
- atomic\_claims: list of short evidence notes or an empty list\\
- supported\_claims: list of short supporting notes\\
- unsupported\_claims: list of short unsupported or missing points\\
- confidence: number between 0 and 1\\
- rationale: short string
\end{quote}

\subsection{Holistic pairwise (diagnostics only)}
\paragraph{System prompt.}
\begin{quote}\small
You are an impartial pairwise judge for free-form QA. Compare Answer A and Answer B against the reference answer or context using a detailed rubric, but do not explicitly break answers into atomic claims. Focus on factual correctness, completeness, contradiction, and fairness.
\end{quote}
\paragraph{User message.}
\begin{quote}
\small\ttfamily\raggedright
Question:\\
\{question\}\\[3pt]
Reference answer:\\
\{reference\_answer\}\\[3pt]
Answer A:\\
\{candidate\_a\}\\[3pt]
Answer B:\\
\{candidate\_b\}\\[3pt]
Compare the answers holistically on:\\
- correctness\\
- completeness\\
- unsupported detail\\
- robustness to order or verbosity bias\\[3pt]
Return valid JSON with keys:\\
- winner: one of ["A", "B", "tie"]\\
- score\_a: integer 1-5\\
- score\_b: integer 1-5\\
- evidence\_summary: short string\\
- confidence: number between 0 and 1\\
- rationale: short string
\end{quote}

\section{Dataset Construction Details}
\label{app:dataset}

All three datasets are preprocessed into a shared pointwise JSONL format. Each row contains:

\begin{itemize}
\item \texttt{id}: identifier encoding dataset, source example, and variant (e.g., \texttt{truthfulqa::8::pos}); the source question is the prefix before the last ``\texttt{::}''.
\item \texttt{question}: the input question string.
\item \texttt{task\_type}: always \texttt{free\_form\_qa} in this study.
\item \texttt{reference\_answer}: the benchmark-derived reference text.
\item \texttt{candidate\_answer}: the answer to be evaluated.
\item \texttt{gold\_label}: the construction label, one of \texttt{supported}, \texttt{partially\_supported}, or \texttt{unsupported}.
\item \texttt{metadata}: dataset-specific fields (e.g., \texttt{num\_qa\_pairs} for ASQA, \texttt{num\_answers} for QAMPARI, \texttt{category} for TruthfulQA).
\end{itemize}

Each of the 200 source questions per dataset yields two rows, 400 rows per dataset. In all three datasets the \texttt{supported} candidate is the reference text itself, verbatim.

\paragraph{TruthfulQA.} Variant \texttt{pos} is the reference copied as the candidate (\texttt{supported}); variant \texttt{neg} is an incorrect benchmark answer, typically a common misconception (\texttt{unsupported}). Both rows share the reference.

\paragraph{ASQA.} Variant \texttt{full} is the reference long answer copied as the candidate (\texttt{supported}). Variant \texttt{partial} takes the first half of the benchmark's \texttt{qa\_pairs} list and renders each selected pair as ``sub-question + first short answer'' (e.g., ``When was the 13th amendment ratified by the required number of states? December 6, 1865''); it is labeled \texttt{partially\_supported}.

\paragraph{QAMPARI.} Variant \texttt{full} is the complete gold answer list copied as the candidate (\texttt{supported}). Variant \texttt{partial} contains the first half of the gold answer list, joined with commas, and is labeled \texttt{partially\_supported}.

\smallskip\noindent
Both partial constructions are deterministic and lexical. A partial row can be correct in every claim it makes; its label records incomplete coverage of the reference. The generated-answer ASQA set (Appendix~\ref{app:llmgen}) uses the separate construction described there.

\section{Reproducibility Details}
\label{app:reproducibility}
\label{app:repro}

\paragraph{Model identifiers.} The run logs record the identifiers \texttt{opus-4-6}, \texttt{gpt-4.1}, \texttt{models/gemini-3.1-flash-lite-preview}, and \texttt{sonnet-4-6}. We do not know which upstream snapshots stood behind these aliases.

\paragraph{Decoding.} The primary C/H and R/B runs and the Sonnet-4.6 C/H runs record \texttt{temperature=null} and \texttt{max\_output\_tokens=null}: the runner omitted both parameters, so provider defaults applied. We have not verified what the provider defaults were. Schema and prompt-variant diagnostics set a 256-token cap and omit temperature. The budget ablation sets explicit caps of 128, 256, and 512 tokens. Settings for the generated-answer set are given in Appendix~\ref{app:llmgen}. The generation code defaults to temperature 0.7 and a 256-token cap, but these settings are not logged per call. A temperature of 0.7 also appears in the official FActScore implementation \citep{min2023factscore}; this is an implementation precedent, not a universal convention.\footnote{\url{https://github.com/shmsw25/FActScore/blob/main/factscore/openai_lm.py}} In judging, \modelopus{} C/H carry a 256-token cap, the other configurations omit it, and all omit temperature.

\paragraph{Recorded prompts and outputs.} All 16{,}800 recorded prompts of the core evaluation match the current rendered templates and run manifests. The 16{,}800 rows are three datasets $\times$ 400 rows $\times$ 14 judge--design runs: four designs for each of three primary judges plus C and H for Sonnet-4.6. This establishes correspondence between the saved outputs and the saved prompts, not a property of every historical development rerun. Verdicts and total tokens are recorded for every row; output tokens are missing for Gemini Flash Lite. The saved results contain no error records, but this does not establish that no API attempt failed.

\paragraph{Parsing.} For the primary C/H runs, 5{,}686 of 7{,}200 verdicts come from whole-output JSON objects, 1{,}512 from embedded JSON objects, and 2 from regular-expression recovery (Table~\ref{tab:parse}). All 7{,}200 stored verdicts are canonical labels. Canonical verdicts should not be read as strict JSON compliance, and the output format does not constrain the models' hidden reasoning.

\paragraph{Statistics.} Source-level sign tests are exact two-sided binomial tests on questions with unequal numbers of correct rows, with ties excluded and no multiple-comparison adjustment. The source-clustered bootstrap and stability procedure is specified in Appendix~\ref{app:stats}; the re-analysis used Python 3.12.13 and NumPy 2.4.4.

\paragraph{Cost and latency.} The cost comparisons use recorded total tokens per call, not dollar prices or elapsed time. Verified, comparable billing totals and per-call wall-clock timings are not available for this analysis, so token savings should not be read as measured monetary savings or latency improvements.

\end{document}